# IndigoVX: Where Human Intelligence Meets AI for Optimal Decision Making

**Kais Dukes**

Hunna Technology

`kais.dukes@hunna.app`

July 21, 2023

## Abstract

This paper defines a new approach for augmenting human intelligence with AI for optimal goal solving. Our proposed AI, Indigo, is an acronym for **I**nformed **N**umerical **D**ecision-making through **I**terative **G**oal-**O**riented optimization. When combined with a human collaborator, we term the joint system IndigoVX, for Virtual eXpert. The system is conceptually simple. We envisage this method being applied to games or business strategies, with the human providing strategic context and the AI offering optimal, data-driven moves. Indigo operates through an iterative feedback loop, harnessing the human expert's contextual knowledge and the AI's data-driven insights to craft and refine strategies towards a well-defined goal. Using a quantified three-score schema, this hybridization allows the combined team to evaluate strategies and refine their plan, while adapting to challenges and changes in real-time.

## 1. Introduction

Feedback loops are a universal feature of artificial intelligence, underpinning well-established techniques such as gradient descent, genetic algorithms, and reinforcement learning, among others (Goodfellow et al, 2016; Russell and Norvig, 2016). These iterative processes leverage a continuous cycle of adjustment and refinement to guide a system towards a desired outcome (Domingos, 2015). We introduce a novel approach that augments a human expert with an AI to achieve an optimal plan. Our approach is widely applicable, for example to collaborative business strategies, game playing, creative writing, or even planning a route for a road trip.

We consider the planning problem as a classic multi-objective optimization problem, given weights and scores (Deb, 2001). The scores reflect the fitness of a candidate plan for a goal, and the weights reflect the relative importance of each scoring function:

$$argmax \sum_i w_i s_i$$

In the field of automated or machine learning-assisted decision-making, the essential aspect of uncertainty is often overlooked (Kochenderfer, 2015). In Indigo, working together as a collaborative team, the human expert provides context-specific expertise and judgment, while the AI provides data-driven insights, scalability, and iterative problem-solving capacity. Human supervision can reduce uncertainty by providing additional context (Holzinger, 2016).

## 2. The IndigoVX Algorithm

This section provides an overview of the step-by-step process of the IndigoVX algorithm. We begin by defining the objective, participants, and other components needed for the task.

The algorithm itself is simple:

> **Objective:** The goal must be well-defined, specific, and quantifiable.
>
> **Participants:** An AI and a human expert (or an ensemble).
>
> **Initialization step:** Both the AI and the human expert draft an initial plan together. This acts as the baseline for iterations.
>
> **Scoring schema:** Using the initial draft plan and the goal as context, define a scoring system of three scores, each of which is quantized on a 0.5 scale from 0 to 10. This is a set of evaluation criteria. These scores are used to assess the quality and effectiveness of the current plan in achieving the goal.
>
> **Optimization loop:** Jointly, the AI and the human expert proceed with the following steps:
>
> 1. Rate the current plan against the agreed 3-score schema.
> 2. Suggest a list of concrete edits to the plan, and explain why each edit boosts the scores.
>
> **Convergence:** Continue the cycle until the convergence criterion indicates an optimal plan has been reached.

## 3. Inspiration from Reinforcement Learning

IndigoVX is inspired by Reinforcement Learning (RL). Although the AI may not have direct access to the environment, the human expert will, and so is a proxy for the environment (Christiano et al, 2017). In RL, a policy determines the agent's action at a given state (Sutton and Barto, 2018). The iterative loop of scoring and making adjustments to the plan therefore resembles a policy iteration. In Indigo, the policy can be thought of as the strategy to improve the plan based on the expert's feedback and AI's suggestions.

However, Reinforcement Learning typically operates within a mathematically well-defined environment, with states, actions, and rewards. In contrast, IndigoVX operates in a broader, more abstract context. For example, it is possible to extract numerical scores from large-language models (LLMs), and other similar AI systems used as constituent components for IndigoVX, for problems that are not necessarily precisely mathematically defined.

The benefits of this reinforcement learning-inspired approach include:

1. A hybrid intelligence of combined AI and human experts.
2. Integration of complex knowledge and background context.
3. Does not require a detailed mathematical model of the environment.

## 4. Addressing Challenges

While implementing IndigoVX, we encountered several challenges beyond the most common found in classic optimization (Boyd and Vandenberghe, 2004). We offer below some potential solutions on how to overcome these.

**Scoring schema sensitivity:** To mitigate the impact of scoring schema selection, we allow for flexibility in the scoring system. Different types of problems are best modeled using different scoring schemes. The scoring system can also be made adaptive to align with the participants' perspectives and preferences.

**Subjectivity:** We acknowledge that the choice of scores can be considered partly subjective. To address this subjectivity, we find that scoring schemas for similar use cases are best standardized. However, given the complex nature of integration of human and AI knowledge, we argue that partially subjective criteria may be a strength. Some subjectivity helps understanding the human perspective, and provides a more comprehensive approach to problem-solving (March, 1994).

**Convergence:** We implement a strategy loosely inspired by the gradient-descent method traditionally used in the training of neural networks. We define convergence as the point at which the weighted score difference across all three scores from one iteration to the next falls below a defined threshold (for example, 0.5) over a window of several previous iterations. This method allows us to determine when subsequent iterations are no longer resulting in significant improvements to the plan, indicating that an optimal or near-optimal plan has been reached. It is expected that each iteration of the feedback loop intuitively follows a path loosely equivalent to steepest descent.

## 5. Scoring Scale

For scoring functions, we select a quantized 0.5 scale from 0 to 10, for three main reasons.

1. **Familiarity:** Scoring out of 10 is widely used, such as in surveys, movie ratings, and psychometric studies.

2. **Cognitive accessibility:** A scale from 0 to 10, with 0.5 increments, is intuitive and requires less cognitive effort as opposed to scales with a larger range like a 100-point scale.

3. **Midpoint for neutral positions:** As opposed to a scale from 1 to 10, a 0 to 10 scale provides a clear midpoint (5). This enables the human expert to assign an intermediate score when they have no strong opinions about a certain part or aspect of the plan.

Incorporating 0.5 increments offers more flexibility. From our trials, we observed that human experts might hesitate when asked to choose between two consecutive whole numbers, such as 7 or 8. The option of selecting 7.5 can resolve this indecision and enable more nuanced scoring, enhancing the precision of evaluations without overwhelming the participant.

## 6. Dynamic Weights

We model IndigoVX as an optimization problem in parameter space, with the scoring functions providing a measure of the plan fitness. The weights assigned to plan scores can be dynamic, not static. They can be adjusted as part of the feedback loop if the expert feels some aspects of their task need more focus. A key question we addressed was how to combine the three scoring functions into a single measure, considering the multi-objective nature of the optimization problem at hand. Inspired by approaches to Pareto optimization, we assign a weighted sum to normalized scores. The weights reflect the importance or priority of each score in the particular context of the plan.

To ensure the process is as adaptable as possible, the weights assigned to each score can be adjusted over time. This is beneficial if the situation changes or if the participant finds that certain aspects of the plan are more important than initially believed (Miettinen and Mäkelä, 2006). The weights can be updated based on the feedback from the human expert, making the process collaborative and dynamic.

In multi-objective optimization, traditional weighted sum approaches can bias solutions towards the corners of the Pareto front (Marler and Arora, 2004). However, our dynamic weighting system overcomes this by continually adjusting the importance of each objective based on real-time expert feedback, resulting in balanced solutions. Further refinement using Pareto-based methods, which can ensure a diverse spread of high-quality solutions across the Pareto front, remains an area for future exploration.

# 7. Collective Intelligence

A collective intelligence (Malone, 2018), also known as a hive mind, can potentially tackle more complex optimization problems through the 'wisdom of crowds' (Surowiecki, 2004; Woolley et al, 2010). For instance, it's commonly understood that a single person's estimate of the number of coins in a jar is unlikely to be accurate. However, surprisingly, combining estimates from multiple people often yields a more accurate result.

There are many possible choices for the architecture of such an ensemble network. For example, multiple human experts could be teamed with a single AI, or a single human expert could be teamed with multiple AI systems. In a practical setting, we have had the most success combining a single human expert with only the most capable systems presently in use. These include Open AI ChatGPT 4, Google Bard and more recently Anthropic Claude 2. Although more computationally expensive, one straightforward extension to multiple AIs is to introduce an additional voting step. In this extension, multiple participants in the ensemble each independently suggest edits to improve the plan scores, and then all participants, including human experts, vote on the next move through parameter space.

# 8. Relationship to Traditional Decision-Making Matrices

The choice of three scoring factors is designed to balance the need for comprehensive problem-solving without overwhelming cognitive load. While the algorithm does not explicitly mimic traditional decision-making matrices, it is capable of capturing similar dynamics when appropriately parameterized, for example:

- Ease and Effect Matrix: This matrix takes into account the impact and effort (reciprocal of ease) required for each task (Boardman, et al 2018).

- Iron Triangle: Often used in project management, this matrix considers the speed, quality, and cost of a project (Atkinson, 1999).

- Eisenhower Matrix: This matrix categorizes tasks based on their importance and urgency (Covey, 2020).

Each factor in these traditional matrices could be represented as one of the three scores, effectively translating qualitative decision-making processes into a quantified format. Integration with AI facilitates a degree of scalability and data-driven decision-making that enhances the precision and effectiveness of the optimization process.

## 9. Conclusion

This decade has arguably already seen the emergence of early-stage Artificial General Intelligence (often called proto-AGI), which can understand, learn from, and apply knowledge across a broad spectrum of tasks. We envisage a future where a select portion of humanity may opt for physical neural augmentation with AI, a process where invasive artificial components enhance human neurology through physical Brain-Computer Interfaces (BCI), such as implants or neural laces.

In this potential scenario, Indigo might be viewed as a relatively simplistic, primitive approach. However, Indigo provides a foundational basis for human-AI collaborative problem-solving, and can be seen as an alternative first step towards non-invasive human-AI hybrid intelligence. The algorithm is simple to implement, arguably requiring only a pen and paper, and access to high-quality natural language AI models.